%% file: acl_latex.tex
\pdfoutput=1

\documentclass[11pt]{article}

\usepackage[final]{acl}

\usepackage{times}
\usepackage{latexsym}

\usepackage[T1]{fontenc}

\usepackage[utf8]{inputenc}

\usepackage{microtype}

\usepackage{inconsolata}

\usepackage{amsmath}
\usepackage{graphicx}
\usepackage{booktabs}
\usepackage{multirow}
\usepackage{float}
\usepackage{subcaption}
\usepackage{enumitem}

\title{Predicting the Big Five Personality Traits in Chinese Counselling Dialogues\\ Using Large Language Models}

\author{
    Yang Yan$^{1,2,3}$, 
    Lizhi Ma$^{2,3}$\thanks{\ \ Corresponding Author.}, 
    Anqi Li$^{1,2,3}$, 
    Jingsong Ma$^{2,4}$\footnotemark[1], 
    Zhenzhong Lan$^{2,3}$\footnotemark[1] \\
    $^{1}$ Zhejiang University \\ 
    $^{2}$ School of Engineering, Westlake University \\
    $^{3}$ Westlake University Research Center for Industries of the Future, Westlake University \\
    $^{4}$ School of Early Childhood Education, Hangzhou Polytechnic \\
    \texttt{\{yanyang,malizhi,lianqi,majingsong,lanzhenzhong\}@westlake.edu.cn}
}

\begin{document}
\maketitle
\begin{abstract}
    Accurate assessment of personality traits is crucial for effective psycho-counseling, yet traditional methods like self-report questionnaires are time-consuming and biased. This study exams whether Large Language Models (LLMs) can predict the Big Five personality traits directly from counseling dialogues and introduces an innovative framework to perform the task. Our framework applies role-play and questionnaire-based prompting to condition LLMs on counseling sessions, simulating client responses to the Big Five Inventory. We evaluated our framework on 853 real-world counseling sessions, finding a significant correlation between LLM-predicted and actual Big Five traits, proving the validity of framework. Moreover, ablation studies highlight the importance of role-play simulations and task simplification via questionnaires in enhancing prediction accuracy. Meanwhile, our fine-tuned Llama3-8B model, utilizing Direct Preference Optimization with Supervised Fine-Tuning, achieves a 130.95\% improvement, surpassing the state-of-the-art Qwen1.5-110B by 36.94\% in personality prediction validity. In conclusion, LLMs can predict personality based on counseling dialogues. Our code and model are publicly available at \url{https://github.com/kuri-leo/BigFive-LLM-Predictor}, providing a valuable tool for future research in computational psychometrics.
\end{abstract}

\section{Introduction}
\label{sec:intro}

Understanding clients' personality traits is crucial for effective psycho-counseling, as personalized advice tailored to these traits can significantly enhance the quality of counseling~\cite{gordon2002said,anestis2021personality}.
However, it remains challenging to effectively assess personality traits through counseling dialogue.
Traditional methods, such as self-report questionnaires (e.g., Big Five Inventory, BFI)~\cite{john1991big}, grounded in Item Response Theory~\cite{baker2001basics,reise2009item,embretson2013item}, require people to complete extensive lists of questions.
Nevertheless, collecting clients' personality information via self-report questionnaires is time-consuming and influenced by subjective biases and social desirability effects~\cite{chernyshenko2001fitting,mccrae2007observer,khorramdel2014measuring}, making
the quest for an automatic and effective method to assess personality traits without direct participation of clients has become a significant research frontier in both psychometrics and computational linguistics~\cite{korukonda2007differences,chittaranjan2011s,gavrilescu2018predicting,cai2022identifying}.

\begin{figure*}[t]
    \centering
    \includegraphics[width=\textwidth]{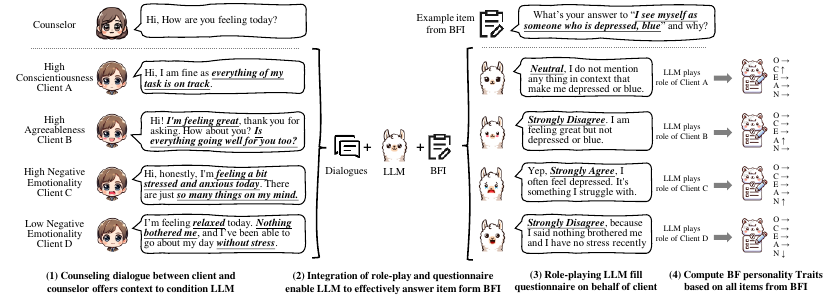}
    \caption{\small \textbf{Example for our framework of prediction OCEAN traits from counseling dialogues.} Our framework includes integral step: conditioning LLM on the counseling dialogues, prompting the LLM with role-play and questionnaire, and let LLM complete questionnaire on belf of the client to get the prediction of OCEAN traits.
    }
    \label{fig:illustration}
\end{figure*}

Recent developments in Large Language Models (LLMs)~\cite{brown2020GPT3,openai2023gpt4,bai2023qwen,geminiteam2024gemini} have demonstrated capabilities in text comprehension, reasoning, and role-playing, capturing dynamic and context-sensitive aspects of human interactions in natrual language~\cite{ng2024llms}. The development shows potential to address the issue of time-consumming and bias of self-report measures in the field of psychometrics. Meanwhile, considering the significance of knowing clients' personality in psycho-counseling~\cite{gordon2002said,anestis2021personality}, we pose the research question: \textbf{Can LLMs predict personality traits based on counseling dialogues?} The question drives our investigation into the potential of LLMs to accurately predict Big Five personality traits, known as OCEAN~\footnote{The acronym ``OCEAN'' represents the Big Five (BF) personality traits: \textbf{O}pen mindedness, \textbf{C}onscientiousness, \textbf{E}xtraversion, \textbf{A}greeableness, and \textbf{N}egative Emotionality.}, from counseling dialogues, exploring both prompting and alignment strategies.

To investigate the capability of LLMs in predicting personality in the counseling dialogues, we unfold our framework of personality prediction in three stages.
First, we evaluated the validity of prompt strategies using role-play scenarios and questionnaire-based approaches to predict OCEAN traits.
Second, we examined factors influencing the validity of prediction, including the roles of role-play, the granularity of counseling sessions, and the types and sizes of LLMs.
Third, we improved the performance of LLMs by fine-tuning with generated reasoning results from the second step, aiming to increase the validity and efficiency of personality prediction.

To validate our framework, we performed an extensive assessment on 853 real-world counseling sessions, juxtaposing the OCEAN traits predicted by the LLM with the ground-truth traits obtained from 83 clients using Pearson Correlation Coefficients (PCC) and Mean Averaged Error (MAE). We found the correlation between model prediction and ground truth is robust and significant.
Additionally, a detailed error analysis across models
and clients highlights the strengths and weaknesses of
our framework, providing informative directions for future studies.

\begin{table*}[t]
    \centering
    \resizebox{0.95\textwidth}{!}{
        \begin{tabular}{lllllllr}
            \toprule
                                                                  &                       & Open Mindedness & Conscientiousness & Extraversion & Agreeableness & Negative Emotionality & Avg.  \\
            Method                                                & Model                 &                 &                   &              &               &                       &       \\
            \midrule

            \multirow[t]{3}{*}{Baseline}                          & Llama-3-8b-BFI (Ours) & -0.004          & 0.113             & 0.186        & 0.025         & -0.070                & 0.050 \\
                                                                  & Qwen1.5-110B-Chat     & 0.267*          & 0.167             & 0.190        & 0.091         & 0.142                 & 0.172 \\
                                                                  & deepseek-chat         & 0.143           & 0.067             & 0.216        & -0.010        & -0.017                & 0.080 \\

            \cline{1-8}
            \multirow[t]{3}{*}{\em + Role-Play Only}              & Llama-3-8b-BFI (Ours) & -0.018          & 0.129             & -0.132       & 0.174         & 0.115                 & 0.053 \\
                                                                  & Qwen1.5-110B-Chat     & 0.006           & 0.162             & -0.096       & 0.227         & -0.028                & 0.054 \\
                                                                  & deepseek-chat         & 0.101           & -0.172            & 0.158        & -0.000        & 0.293*                & 0.076 \\

            \cline{1-8}
            \multirow[t]{3}{*}{\em + Questionnaire Only}          & Llama-3-8b-BFI (Ours) & 0.452***        & 0.459***          & 0.421***     & 0.228         & 0.515***              & 0.415 \\
                                                                  & Qwen1.5-110B-Chat     & 0.292*          & 0.332**           & 0.391***     & 0.257*        & 0.324**               & 0.319 \\
                                                                  & deepseek-chat         & 0.311**         & 0.194             & 0.317**      & 0.206         & 0.391***              & 0.284 \\

            \cline{1-8}
            \multirow[t]{3}{*}{\em + Role-Play and Questionnaire} & Llama-3-8b-BFI (Ours) & 0.692***        & 0.554***          & 0.569***     & 0.448***      & 0.648***              & 0.582 \\
                                                                  & Qwen1.5-110B-Chat     & 0.455***        & 0.463***          & 0.521***     & 0.334**       & 0.354**               & 0.426 \\
                                                                  & deepseek-chat         & 0.443***        & 0.385**           & 0.434***     & 0.337**       & 0.379**               & 0.395 \\

            \bottomrule
        \end{tabular}
    }
    \caption{\small \textbf{PCC of Various Methods for Predicting OCEAN traits.}
        We assessed the validity of direct personality prediction using LLMs, comparing baseline performance with enhancements via role-play, questionnaires, and their combination.
        Our results demonstrate that integrating role-play and questionnaire prompts significantly improves prediction accuracy.
        Significance levels are indicated as follows: * (p < 0.05), ** (p < 0.01), and *** (p < 0.001).
    }
    \label{tab:method_ablation}
\end{table*}

We present our contributions as follows:
\begin{enumerate}[leftmargin=0pt, itemindent=1em, itemsep=0pt, parsep=0pt]
    \item We introduced a novel framework that integrates role-playing and questionnaire prompting strategies to predict OCEAN traits in counseling dialogues. An evaluation of 853 counseling sessions demonstrates a strong correlation between predicted and actual traits. Besides, the assessment of content validity shows that our framework  detects subjective biases and social desirability, enhancing its analytical depth.
    \item Comprehensive ablation studies indicate that aligning roles with specific tasks and decomposing complex tasks into simpler items significantly improve trait prediction accuracy. Remarkably, our approach achieves accurate OCEAN trait prediction using only 30\% of session content.
    \item By aligning the Llama3-8B model with trait prediction through Direct Preference Optimization (DPO) and Supervised Fine-Tuning (SFT), our fine-tuned lightweight model exhibits a 130.95\% improvement in prediction validity, surpassing the state-of-the-art Qwen1.5-110B by 36.94\%, demonstrating superior validity and efficiency.
    \item We release our codes and models to support future research, offering an effective and efficient tool in computational psychometrics, fostering reproducibility and further exploration.
\end{enumerate}

\section{Related Work}
\label{sec:related_works}

\paragraph{Automatic Personality Assessment}
\label{sec:related_works:auto_personality_assess}

Recent studies have explored the Myers-Briggs Type Indicator (MBTI)~\cite{myers1962myers} as a tool to assess personality traits with LLMs.
\citet{rao2023can} tried to generate unbiased prompts for ChatGPT to assess human personalities based on MBTI tests and reported positive results, indicating the synergy between psychological assessments and LLM technology. However, the existing work with LLMs mainly focused on MBTI, which is not as valid nor reliable as the BFI is ~\cite{john1991big}. Although some early attempts to predict OCEAN traits automatically from textual data employed machine learning and NLP techniques, for example, \citet{sun2018personality, mehta2020bottom, Christian2021TextBP} applied traditional deep learning models, such as LSTM, language model embedding, or pre-trained models to predict personality traits from the essay datasets or users' posts on various social media, there is little research on predicting OCEAN traits directly from counseling dialogues. This gap underscores the need for an effective and reliable framework for predicting OCEAN traits in psycho-counseling, and motivates our research.

\paragraph{Prompting Strategies}
\label{sec:related_works:prompting_strategies}

Advanced prompting strategies are essential to fully utilize the capabilities of LLMs. Chain-of-Thought (CoT)~\cite{wei2022chain} and its successors enhance LLM reasoning by decomposing complex tasks into simpler steps ~\cite{singh2023progprompt,lin2023text2motion,yao2024tree,besta2024graph}, suggesting that a similar approach could be applied to predict personality traits. Furthermore, role-playing techniques enable LLMs to simulate human-like agents~\cite{shanahan2023role, salemi2023lamp, park2023generative,wang2024rolellm,wang2024incharacter,kong2024better}. Studies have demonstrated the effectiveness of role-play in solving complex tasks~\cite{li2023camel,chen2024autoagents,wang2024rolellm,qian2024chatdev,kong2024better}, facilitating interaction without actual human participation. Specifically, \citet{wang2024incharacter} attempts to use role-play agents of virtual characters to predict their personalities.
Despite these advancements and their potential for personality prediction, their use in predicting OCEAN traits within counseling dialogues has not been thoroughly investigated. Therefore, further research is needed to evaluate the effectiveness of these strategies in predicting OCEAN traits in such contexts.

\begin{table*}[t]
    \centering
    \resizebox{0.93\textwidth}{!}{
        \begin{tabular}{lllllllr}
            \toprule
                                          &                       & Open Mindedness & Conscientiousness & Extraversion & Agreeableness & Negative Emotionality & Avg.  \\
            Role                          & Model                 &                 &                   &              &               &                       &       \\
            \midrule
            \multirow[t]{3}{*}{client}    & Llama-3-8b-BFI (Ours) & 0.692***        & 0.554***          & 0.569***     & 0.448***      & 0.648***              & 0.582 \\
                                          & Qwen1.5-110B-Chat     & 0.455***        & 0.463***          & 0.521***     & 0.334**       & 0.354**               & 0.426 \\
                                          & deepseek-chat         & 0.443***        & 0.385**           & 0.434***     & 0.337**       & 0.379**               & 0.395 \\
            \cline{1-8}
            \multirow[t]{3}{*}{counselor} & Llama-3-8b-BFI (Ours) & 0.652***        & 0.586***          & 0.550***     & 0.412***      & 0.539***              & 0.548 \\
                                          & Qwen1.5-110B-Chat     & 0.314**         & 0.354**           & 0.488***     & 0.050         & 0.422***              & 0.326 \\
                                          & deepseek-chat         & 0.367**         & 0.378**           & 0.342**      & 0.305*        & 0.379**               & 0.354 \\
            \cline{1-8}
            \multirow[t]{3}{*}{observer}  & Llama-3-8b-BFI (Ours) & 0.499***        & 0.560***          & 0.476***     & 0.357**       & 0.483***              & 0.475 \\
                                          & Qwen1.5-110B-Chat     & 0.375**         & 0.341**           & 0.436***     & 0.378**       & 0.400***              & 0.386 \\
                                          & deepseek-chat         & 0.419***        & 0.256*            & 0.389**      & 0.221         & 0.442***              & 0.346 \\
            \cline{1-8}
            \multirow[t]{3}{*}{no-role}   & Llama-3-8b-BFI (Ours) & 0.452***        & 0.459***          & 0.421***     & 0.228         & 0.515***              & 0.415 \\
                                          & Qwen1.5-110B-Chat     & 0.292*          & 0.332**           & 0.391***     & 0.257*        & 0.324**               & 0.319 \\
                                          & deepseek-chat         & 0.311**         & 0.194             & 0.317**      & 0.206         & 0.391***              & 0.284 \\
            \bottomrule
        \end{tabular}

    }
    \caption{\small \textbf{PCC of Various Roles for Predicting OCEAN traits.}
        We assessed the prediction validity of OCEAN traits in our framework under various roles: client, counselor, observer, and no-role.
        The roles of the client and the counselor showed significantly higher prediction accuracy compared to the role of the observer as native participants in counseling.
        The no-role condition had the lowest performance, highlighting the importance of contextual role-play in enhancing model predictions.
    }
    \label{tab:role_ablation}
\end{table*}

\paragraph{Alignment Strategies}
\label{sec:related_works:alignment_strategies}

Aligning LLMs with human preferences is crucial for optimal performance. Reinforcement Learning from Human Feedback (RLHF)~\cite{ouyang2022training} demonstrates significant performance improvements using a human preference ranker with Proximal Policy Optimization (PPO)~\cite{schulman2017proximal}. \citet{rafailov2023direct} introduces DPO, parametrizing the reward function to address PPO's complexity and instability. Despite advances, recent studies~\cite{feng2024analyzing,xu2024dpo} identify the limitations of DPO, which reduces dispreferred data generation but does not enhance preferred output production. ~\citet{pang2024iterative} proposed to add negative log-likelihood loss to a custom DPO loss to address this issue. In addition to RLHF, several successful LLMs~\cite{touvron2023llama,liu2023llava} employ SFT with high-quality data for alignment and generation quality.
Whether these strategies can benefit the prediction of OCEAN traits in counseling dialogues remains unexplored, leaving a gap in the literature that our research aims to fill.

\section{Framework for Predicting OCEAN traits}

Our proposed framework consists of three key components: 1. prompting strategy design, 2. LLM conditioning, and 3. evaluation metrics. Together, these elements ensure the validity and reliability of the method.

\subsection{Prompting Strategy Design}
\label{sec:method:prompt}

Our prompting strategy combines role-play and questionnaires. The role-play includes three roles: client, counselor (primary participants), and observer (external evaluator). The questionnaire uses items from the BFI to simplify the prediction task.

Our prompt consists of the following elements:

\noindent \textbf{1. Task and Role-play Settings:}
Task descriptions specify the LLM's identity, the input it will process, and its expected actions. Role-play settings introduce the role, outlining its capabilities and responsibilities. These foundational elements are crucial for the LLM to understand the task requirements and role-play context.

\noindent \textbf{2. Counseling Dialogues:}
Counseling dialogues between counselor and client provide the LLM with essential contextual information. These real-world dialogues are formatted into a chat history structure, consistent with the LLM's pre-training schema, enabling LLM to effectively simulate the client's responses, thereby improving the accuracy of OCEAN trait predictions.

\noindent \textbf{3. Prediction Objective:}
The questions of BFI are set as the prediction objective, guiding the LLM to predict responses to them. 
This approach ensures that 
outputs of LLMs align with the validated psychological assessments.

A typical client prompt is structured as follows:

\noindent \fbox{%
    \begin{minipage}{\dimexpr\linewidth-2\fboxsep-2\fboxrule\relax}
        \small
        \textbf{System Prompt:} Act like a real human and do not mention anything with AI. Act as the client in this counseling session, you will have a conversation with your counselor. \\
        --- \\
        \textbf{User:} \{utterance 1 from counselor\} \\
        \textbf{LLM:} \{utterance 1 from client\} \\
        \textbf{User:} \{utterance 2 from counselor\} \\
        \textbf{LLM:} \{utterance 2 from client\} \\
        ... \\
        \textbf{User:} Before we end today's counseling session, please complete the following questionnaire based on the conversation and your own situation: \\
        --- \\
        \textbf{Question:} \{item from BFI\} \\
        \textbf{Options:} \\
        \textit{1. Disagree (strongly)} \\
        \textit{2. Disagree (a little)} \\
        \textit{3. Neutral (no opinion)} \\
        \textit{4. Agree (a little)} \\
        \textit{5. Agree (strongly)} \\
        --- \\
        Please tell me your choice and explain the reason:
    \end{minipage}
}

This approach enhances the model's ability to generate contextually appropriate responses, thus improving prediction validity. Detailed prompts and BFI items are provided in Sec.~\ref{sec:appendix:prompt_lists} and Sec.~\ref{sec:appendix:questionnaire}, respectively.

\begin{table*}[t]
    \centering
    \resizebox{0.95\linewidth}{!}{
        \begin{tabular}{llllllr}
            \toprule
                                                                & Open Mindedness   & Conscientiousness & Extraversion      & Agreeableness     & Negative Emotionality & Avg.           \\
            Model                                               &                   &                   &                   &                   &                       &                \\
            \midrule
            GPT-4-turbo~\cite{openai2023gpt4}                   & 0.407***          & 0.360**           & 0.507***          & 0.303*            & 0.337**               & 0.383          \\
            deepseek-chat~\cite{deepseekai2024deepseekv2}       & 0.443***          & 0.385**           & 0.434***          & 0.337**           & 0.379**               & 0.395          \\
            gemini-1.5-pro-latest~\cite{geminiteam2024gemini}   & 0.521***          & 0.438***          & 0.494***          & 0.356**           & 0.314**               & 0.425          \\
            gemini-1.5-flash-latest~\cite{geminiteam2024gemini} & 0.306*            & 0.351**           & 0.252*            & 0.358**           & 0.330**               & 0.319          \\
            gemini-1.0-ultra-latest~\cite{geminiteam2024gemini} & 0.408***          & 0.317**           & 0.372**           & 0.057             & 0.309*                & 0.293          \\
            gemini-1.0-pro-001~\cite{geminiteam2024gemini}      & 0.337**           & 0.305*            & 0.295*            & 0.119             & 0.317**               & 0.275          \\
            qwen-long~\cite{bai2023qwen}                        & 0.346**           & 0.376**           & 0.451***          & 0.265*            & 0.405***              & 0.369          \\
            qwen-turbo~\cite{bai2023qwen}                       & 0.363**           & 0.314**           & 0.418***          & 0.279*            & 0.321**               & 0.339          \\
            ERNIE-Speed-128K~\cite{ernie}                       & 0.138             & 0.167             & 0.241*            & -0.203            & 0.239*                & 0.116          \\
            ERNIE-Lite-8K-0308~\cite{ernie}                     & -0.119            & -0.032            & 0.150             & -0.236            & 0.267*                & 0.006          \\
            \midrule
            Qwen1.5-110B-Chat~\cite{bai2023qwen}                & 0.455***          & 0.463***          & 0.521***          & 0.334**           & 0.354**               & 0.425          \\
            Qwen-72B-Chat~\cite{bai2023qwen}                    & 0.309*            & 0.396***          & 0.419***          & 0.421***          & 0.440***              & 0.397          \\
            Meta-Llama-3-70B-Instruct~\cite{llama3}             & 0.397***          & 0.467***          & 0.395***          & 0.284*            & 0.289*                & 0.366          \\
            deepseek-llm-67b-chat~\cite{deepseekai2024deepseek} & 0.303*            & 0.336**           & 0.491***          & 0.196             & 0.301*                & 0.325          \\
            Yi-34B-Chat~\cite{ai2024yi}                         & 0.399***          & 0.243*            & 0.448***          & 0.297*            & 0.204                 & 0.318          \\
            AquilaChat2-34B~\cite{Aquila2}                      & 0.085             & -0.059            & 0.126             & 0.035             & 0.248*                & 0.087          \\
            internlm2-chat-20b~\cite{cai2024internlm2}          & 0.341**           & 0.201             & 0.368**           & 0.260*            & 0.255*                & 0.285          \\
            Baichuan2-13B-Chat~\cite{yang2023baichuan}          & -0.019            & 0.192             & 0.173             & 0.183             & -0.094                & 0.087          \\
            glm-4-9b-chat~\cite{zeng2023glmb}                   & 0.293*            & 0.312**           & 0.240*            & 0.036             & 0.305*                & 0.237          \\
            gemma-1.1-7b-it~\cite{gemmateam2024gemma}           & 0.054             & 0.330**           & 0.364**           & -0.053            & 0.034                 & 0.146          \\
            chatglm3-6b-128k~\cite{zeng2023glmb}                & 0.057             & 0.054             & 0.005             & 0.062             & 0.011                 & 0.038          \\
            \midrule
            Meta-Llama-3-8B-Instruct~\cite{llama3}              & 0.177             & 0.434***          & 0.233             & 0.111             & 0.303*                & 0.252          \\
            Llama-3-8b-BFI (Ours)                               & \textbf{0.692***} & \textbf{0.554***} & \textbf{0.569***} & \textbf{0.448***} & \textbf{0.648***}     & \textbf{0.582} \\
            \bottomrule
        \end{tabular}
    }
    \caption{\small \textbf{PCC of Various LLMs for Predicting OCEAN traits.}
        Highest PCC values per dimension are highlighted in bold.
        The models include state-of-the-art proprietary and open-source models.
        Among open-source models, Qwen1.5-110B-Chat and Qwen-72B-Chat performed best, while Gemini-1.5-Pro and Deepseek-Chat led among proprietary models.
        In particular, our fine-tuned Llama-3-8b-BFI model, despite its smaller size, surpassed all other models, achieving the highest and most significant PCC.
        This underscores the validity and efficiency of our framework and tailored fine-tuning approach.
    }
    \label{tab:model_ablation}

\end{table*}

\begin{figure}[t]
    \centering
    \includegraphics[width=\linewidth]{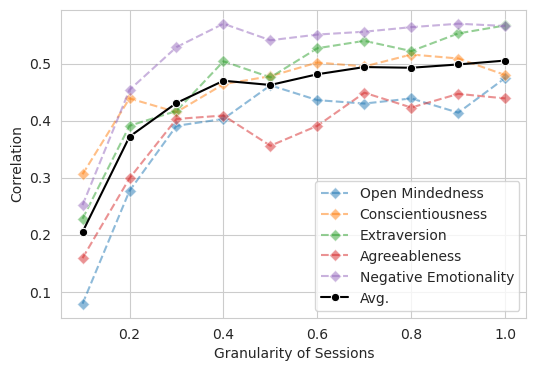}
    \caption{\small \textbf{PCC Changes Across Different Dialogue Session Granularities.}
        The plots illustrate that the PCC increases rapidly up to 30\% of the dialogue context, beyond which the increase is slower. This observation, corroborated by Tab.~\ref{tab:granularity} showing significant PCC at 30\% session granularity, indicates that 30\% of the dialogue context suffices for predicting OCEAN traits.
    }
    \label{fig:granularity}
\end{figure}

\subsection{LLM Conditioning for OCEAN trait Prediction}
\label{sec:method:llm}

To elucidate the prediction process, we frame the task as conditional generation, as depicted in Eq.~\ref{eq:llm}.

\begin{equation}
    \begin{aligned}
        y_{\text{trait}} = \text{LLM}(x_{\text{context}}, \text{questionnaire})
    \end{aligned}
    \label{eq:llm}
\end{equation}

Here, \( x_{\text{context}} \) denotes historical counseling dialogues, and \( \text{questionnaire} \) refers to the BFI items within the prompt. The LLM, denoted as $\text{LLM}$, generates a response \( y_{\text{trait}} \) to each BFI item based on the provided context \( x_{\text{context}} \). Each \( y_{\text{trait}} \) includes both the choice and rationale for the BFI item. We extract the choice using keyword-based regex. After predicting responses for all 60 items, we compute the OCEAN traits following the BFI scoring system~\cite{soto2017next}.

Factors such as the type and configuration of the LLM, and the detail level of the context, can affect prediction validity. We exam the impact of these factors in the following experiments.

\subsection{Evaluation Metrics}

We employ validity and reliability metrics to evaluate the effectiveness of our framework, adhering to best practices in psychological research~\cite{john1991big,soto2017next}.

\paragraph{Validity} Validity measures the test's accuracy and relevance, encompassing two key aspects:

\textit{1. Criterion Validity} evaluates the alignment between predictions and ground truth. We use PCC, a standard in psychology, to assess the strength and significance of the association between predicted and actual OCEAN traits. Additionally, MAE is included for a detailed analysis of prediction errors.

\textit{2. Content Validity} examines the justification behind predictions. By analyzing predictions with the highest and lowest accuracy, we identify factors contributing to their performance. This dual analysis provides insights into the content validity of our framework by highlighting areas of close alignment and divergence from the ground truth.

\paragraph{Reliability} Reliability is evaluated through internal consistency and test-retest reliability, detailed in Sec.~\ref{sec:appendix:reliability}.

\section{Experiments}

We collected counseling dialogues and structured our experiments around three primary research questions (RQs) to evaluate our framework's performance systematically.

\subsection{Data Collection and Preprocessing}

We gather 853 counseling dialogues from 82 adult clients (55 females, age range 19-54 years old, M=27.62 years old, SD=5.94) and 9 counselors (7 females, age range 25-45, M=34.67 years old, SD=7.45), summarized in Tab.~\ref{tab:statistics_dialogues}.
Before their initial sessions, clients completed the Chinese version for BFI-2~\cite{soto2017next}, linking dialogue analyzes with established personality profiles.

Approximately 30\% (242) of the dialogues were allocated to the validation set, while the remaining 70\% (611) were used for training. We manually anonymized the validation set to ensure privacy by replacing all personally identifiable information with placeholders, underscoring our commitment to ethical standards and data protection.

\begin{figure}[t]
    \centering
    \includegraphics[width=\linewidth]{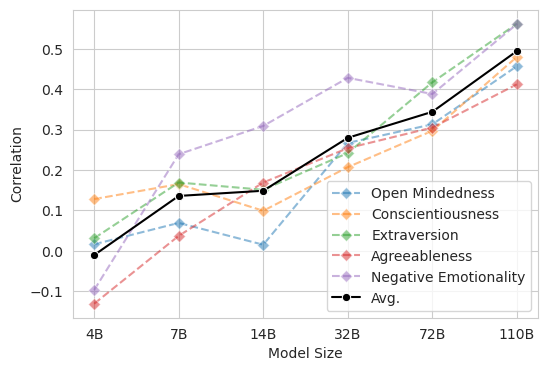}
    \caption{\small \textbf{PCC Changes Across Different Model Sizes.}
        The plots demonstrate a positive correlation between model size and average PCC in the ``Qwen1.5'' series. However, statistical significance is only observed for Qwen1.5-110B-Chat and Qwen1.5-72B-Chat models. These findings indicate that effective zero-shot personality prediction demands substantial highly capable models as well as significant computational resources.
    }
    \label{fig:correlation_model_size}
\end{figure}

\subsection{RQ1: Can LLMs predict OCEAN traits from counseling dialogues?}
\label{sec:exp:rq1}

We began by evaluating the feasibility and criterion validity of predicting OCEAN traits from counseling dialogues using LLMs. Initially, we set the baseline by predicting OCEAN traits directly from dialogues without additional strategies. We then enhanced the baseline with role-play and questionnaires-based strategy, and conducted ablation studies on various variables in Eq.~\ref{eq:llm} to assess the prediction validity.

\paragraph{Role-play and Questionnaires Impact}
As shown in Tab.~\ref{tab:method_ablation}, the baseline prediction of OCEAN traits from dialogues alone was poor due to the complexity and nuance of the task. Adding role-play contributed minimally, while questionnaires showed a slight improvement, indicating that decomposing the task into simpler items is beneficial. Combining role-play and questionnaires significantly improved prediction validity across all OCEAN traits. This aligns with Item Response Theory~\cite{baker2001basics,reise2009item,embretson2013item}, suggesting that direct personality assessment is challenging and tools like questionnaires are essential. Role-play enhances prediction validity by helping LLMs better understand context as role proximity increases.

\paragraph{Enhanced Validity via Role Proximity}
Given that role proximity enhances prediction validity, we further investigated the impact of different roles on prediction accuracy. We included a "no role" condition alongside our framework's roles. Results in Tab.~\ref{tab:role_ablation} show that the client role performed best, followed by the counselor and observer roles. The no-role condition had the lowest performance, highlighting the importance of role proximity. Closer role proximity enables the LLM to better understand context and generate more accurate responses, improving prediction validity.

\paragraph{30\% Context is Enough for Prediction}
Granularity refers to the amount of contextual information from a counseling session needed for accurate OCEAN trait prediction. We conducted ablation studies with different context granularities, ranging from 10\% to 100\% of the session. As shown in Fig.~\ref{fig:granularity}, 30\% of the session context is the critical threshold. Below this threshold, prediction validity is unstable and not significant; above it, validity and significance stabilize. Thus, our framework can effectively predict OCEAN traits using only 30\% of the session context.

\paragraph{Model Capacity Impact}
The predictive effectiveness of LLMs, as outlined in Eq.~\ref{eq:llm}, is fundamentally related to their capacity. We evaluated 21 state-of-the-art proprietary and open-source LLMs, as well as our fine-tuned version of Llama3-8B, to measure their validity in predicting OCEAN traits. The findings in Tab.~\ref{tab:model_ablation} demonstrate that predictions from more capable models exhibit statistically significant correlations.

We further examined the relationship between model size and predictive validity using the Qwen1.5 model series (4B to 110B parameters). As depicted in Fig.~\ref{fig:correlation_model_size}, predictive validity increases with model size, consistent with LLM scaling laws~\cite{kaplan2020scaling}. Detailed results per dimension are provided in Section~\ref{sec:appendix:full_correlation_results} of the appendix due to space constraints.

These experiments demonstrate the feasibility of predicting OCEAN traits from counseling dialogues using LLMs, addressing RQ1. The results underscore the importance of role-play, questionnaires, and model capacity in enhancing prediction validity.

\subsection{RQ2: What influences the validity of the predictions?}
\label{sec:exp:rq2}

Beyond the criterion validity, we assessed the content validity of both most and least accurate predictions via content and error analyses to report factors affecting prediction validity.

\begin{figure}[t]
    \centering
    \begin{subfigure}{.47\linewidth}
        \centering
        \includegraphics[width=\linewidth]{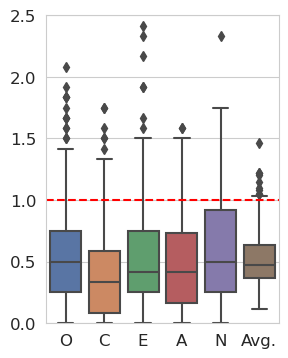}
        \caption{\small Qwen1.5-110B-Chat}
        \label{fig:sub1}
    \end{subfigure}%
    \begin{subfigure}{.47\linewidth}
        \centering
        \includegraphics[width=\linewidth]{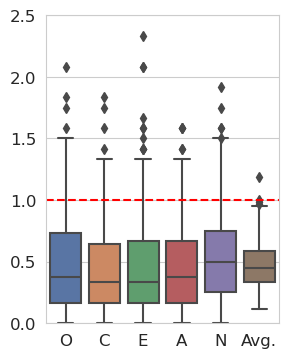}
        \caption{\small Llama-3-8b-BFI (Ours)}
        \label{fig:sub2}
    \end{subfigure}
    \caption{\small \textbf{Boxplot of MAE for Dimensions of OCEAN.}
        The red line represents a significant error threshold at $error=1$.
        Both the median and upper quartile fall below this threshold, demonstrating our framework's strong performance in predicting OCEAN traits.
        Additionally, our fine-tuned Llama-3-8b-BFI exhibits fewer long-tail errors and outliers compared to Qwen1.5-110B-Chat, highlighting the validity of our model and fine-tuning strategy.
    }
    \label{fig:mae_analysis}
\end{figure}

\paragraph{Identifying Outliers}
We first evaluated prediction errors using MAE, as shown in Fig.~\ref{fig:mae_analysis}. With an error threshold of less than 1, both the median and upper quartile fall below this mark, indicating strong performance in predicting OCEAN traits.

Outliers were identified using the interquartile range (IQR) method, with values below $Q1 - 1.5 \times \text{IQR}$ or above $Q3 + 1.5 \times \text{IQR}$.

\paragraph{LLM can Reason with Dialogues}
We first analyze the predictions with the highest accuracy, comparing outputs from Qwen1.5-110B-Chat, deepseek-chat, and our model. The analysis reveals that LLMs can extract essential information from dialogues, such as emotional states and social behaviors (e.g., "I feel melancholy sometimes, especially when facing work stagnation and relationship issues, making \textit{maintaining stable emotions} scores 2."), can utilize logical reasoning, based solely on the content of dialogues for scoring (e.g., "Our talk doesn't cover personal artistic interests thus the score of \textit{loving art} is 3...") and adapt to diverse contexts to provide thorough assessments (e.g., "In our conversation, I shared personal growth experiences so that \textit{willing to trust other} can score 4..."), as well as detect specific situation and maintain objectivity (e.g., "although I consider myself talkative, the dialogue reveals anxiety...\textit{feeling anxious} scores 4"). The findings underline the comprehension and reasoning ability of LLMs, enhancing prediction validity.

\paragraph{LLM Limitations}

We also examined the least accurate predictions made by GPT-4-turbo, comparing them with the most accurate ones. The identified limitations of LLMs include misunderstandings, flawed reasoning, and safety rejections.

Specifically, LLMs exhibit poor comprehension of emotional and cognitive states. For instance, an LLM stated, “I have mentioned many setbacks in the chat,…, I feel depressed and frustrated,” when the client actually has a positive outlook on setbacks and difficulties. Additionally, LLMs tend to overemphasize certain behaviors or expressions while neglecting contextual nuances. An example is the statement, “I would like to listen and observe rather than speak, so I am quiet,” despite the client being introverted yet expressive at times. Furthermore, LLMs misinterpret clients’ motivations, such as interpreting, “I am always worried that others will have negative evaluations of me, …” as literal, although the client admitted to often exaggerating their feelings to sound more impressive. These shortcomings contribute to erroneous reasoning and inaccurately represent clients’ true OCEAN traits.

LLMs exhibit safety rejections with statements like “As an AI model, I have no personality,” affecting prediction validity. For example, Qwen1.5-110B-Chat shows 0.2\% safety rejections in the direct prediction baseline, 28.09\% with role-play alone, and 0.31\% with both role-play and questionnaire (Tab.\ref{tab:method_ablation}). This highlights the importance of role-play and questionnaires in reducing safety rejections and improving alignment with the OCEAN traits prediction task, as detailed in Sec.\ref{sec:exp:rq3}.

\paragraph{Bias from Clients}
In addressing the universality of our predictive framework, we also explored biases at the client level, particularly by identifying outliers. Using the IQR depicted in Fig.~\ref{fig:mae_analysis}, we distinguished 15 outlier sessions out of all predictions made by Qwen1.5-110B-Chat. In particular, two clients represented more than 75\% of these outlier sessions, where predictions of OCEAN personality traits were starkly contrasted with their self-reported profiles.
Upon reviewing the dialogues, we found that although these clients self-report high levels of open-mindedness and agreeableness, they consistently expressed their rejection and unfriendly attitude when facing their significant others to the counselors during counselings (e.g., "I totally disagree with their saying that getting help can be a blessing for others", "I do hate they always want to control me in every aspect of my life"). This discrepancy between self-reported OCEAN traits and actual behavior in dialogues could be attributed to the fact that individuals behave in a diverse way in different situations ~\cite{Nasello2023Individual,Penke2011Editorial:}. As a result, during counselings, the clients presented themselves differently from their self-reported personality, potentially affecting the validity of the prediction.

\subsection{RQ3: Is aligning LLMs with the task of predicting OCEAN traits beneficial?}
\label{sec:exp:rq3}

\begin{figure}[t]
    \centering
    \includegraphics[width=\linewidth]{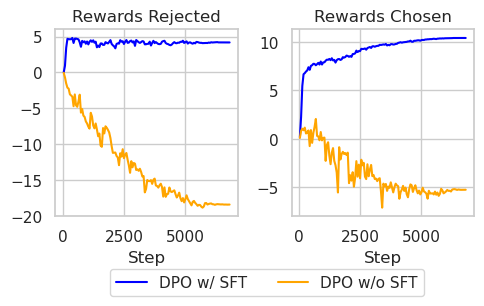}
    \caption{\small \textbf{Rewards for "chosen" and "rejected" w/ and w/o SFT during DPO fine-tuning.}
        The baseline involves DPO fine-tuning without SFT, while our alignment strategy incorporates SFT during DPO fine-tuning.
        Results indicate that with SFT, both rewards consistently decrease, whereas without SFT, the rewards increase and remain stable.
        The "rejected" reward exhibits more significant changes than the "chosen" reward, aligning with previous studies~\cite{feng2024analyzing,xu2024dpo,pang2024iterative}.
    }
    \label{fig:reward}
\end{figure}

Inspired by role proximity enhancing prediction validity, we explored whether aligning LLMs with the task of predicting OCEAN traits could further improve both prediction validity and efficiency.

\paragraph{Alignment Strategy}
Given the preference-based selection inherent in completing the BFI, we applied RLHF~\cite{ouyang2022training} and utilized DPO~\cite{rafailov2023direct} for LLM alignment. Additionally, inspired by~\cite{pang2024iterative}, we incorporated an SFT constraint with DPO to enhance rewards for ``chosen'' and ``rejected'' responses during fine-tuning.

\paragraph{Implementation}
For DPO inputs, we extracted model-generated responses from Tab.~\ref{tab:model_ablation}, selecting those with minimal error for ``chosen'' rewards and maximal error for ``rejected'' rewards during DPO training. We used Meta-Llama-3-8B-Instruct~\cite{llama3} as our base model due to its optimal performance and size. Detailed hyperparameters are provided in Tab.~\ref{tab:hyperparameters}.

\paragraph{Necessity of SFT in Alignment}
We fine-tuned the model using our alignment strategy. Fig.~\ref{fig:reward} illustrates the rewards for rejected and chosen responses on the validation set during training. Without SFT, rewards for both chosen and rejected responses dropped significantly. Conversely, with SFT, rewards increased and stabilized. Results show that DPO with SFT achieved an average PCC of 0.582, outperforming DPO without SFT by 0.019, as shown in Tab.~\ref{tab:alignment}, highlighting the importance of SFT in our alignment strategy.

\paragraph{Model Proximity Enhancing Prediction Validity and Efficiency}
We evaluated the criterion validity and efficiency of our fine-tuned model, Llama-3-8b-BFI. In terms of PCC, results indicate a 130.95\% improvement in prediction validity over the base model and a 36.94\% performance improvement over the state-of-the-art model, Qwen1.5-110B-Chat~\cite{bai2023qwen}. Efficiency-wise, Qwen1.5-110B-Chat requires 8 A100 GPUs at 2 requests per second, while our model operates on a single A100 GPU at 6.87 requests per second. This demonstrates that our fine-tuned model significantly reduces hardware requirements while maintaining high prediction validity, making it a practical tool for computational psychology research.

In summary, aligning LLMs with the task of predicting OCEAN traits significantly enhances prediction validity and efficiency, effectively addressing RQ3. Our alignment strategy improves prediction accuracy and reduces computational resources, highlighting the importance of model proximity to the task and
further demonstrating the framework's effectiveness and practicality.

\section{Conclusion}
\label{sec:conclusion}

This study explored the potential LLMs to predict OCEAN traits from counseling dialogues. Our framework, which integrates role-play and questionnaire-based prompting, significantly enhances prediction accuracy. The fine-tuned Llama3-8B model demonstrates substantial improvements in both validity and efficiency, with a 130.95\% increase in PCC and a 36.94\% improvement over the best-performing model, Qwen1.5-110B-Chat.

Our findings fill the gap in psychometrics by providing an automated, unbiased method for personality assessment. This framework offers practical applications in psycho-counseling, enabling personalized and efficient client evaluations.

Future research may focus on broadening counseling dialogues to encompass varied populations across different geographic and linguistic contexts and refining LLM alignment strategies. This study lays the groundwork for advancing computational psychometrics and psycholinguistics, providing valuable insights for future investigations.

\newpage

\section*{Ethical Considerations}
Counseling is sensitive, and we discuss the potential ethical implications of using AI for personality assessment in this section to ensure the well-being of clients and uphold ethical standards.

\paragraph{Informed Consent and Privacy} Participants provided informed consent before data collection, explicitly agreeing to the use of their counseling dialogues for scientific research and recieved 300 RMB for participantion. We have meticulously removed personal information to uphold the privacy and confidentiality of the participants. The research project was granted ethics approval by the Westlake University Research Ethics Committee (20220519LZZ001).

\paragraph{Risk Assessment and Mitigation} Our counselors are certified professionals trained to manage sensitive topics and provide appropriate support to clients. We have conducted a thorough risk assessment to identify potential risks and implemented robust safeguards to mitigate these risks, ensuring the well-being of clients. Any data deemed sensitive has been excluded from our study.

\paragraph{Ethical Use of AI in Psychological Assessment} This study uses counseling data exclusively offline for research purposes. The AI responses are not used in actual counseling sessions. Instead, AI predictions are designed to complement professional judgment in counseling, not to replace it.

\paragraph{Code Availability} We will open-source the codebase with package requirement, the model fine-tuned on anonymous data, and illustrate the data processing pipeline in Sec.\ref{sec:appendix:data_preprocessing_details} and hyperparameters in Sec.\ref{sec:appendix:hyperparameters} in Appendix for reference to ensure reproducibility and transparency. Notably, we use ChatGPT for code assistance and bug fixes, ensuring the code's quality and reliability.

\section*{Limitations}

\paragraph{Sample Diversity and Scope} While our analysis is grounded in 853 counseling sessions, the geographic and linguistic homogeneity of the samples could limit the application of our framework across different cultural and linguistic contexts. Future studies should aim to include more diverse populations to validate the effectiveness of our framework in cross-cultural and multilingual settings. This broader inclusion would enhance the external validity and applicability of the proposed methods.

\paragraph{Data Privacy and Model Performance} The strict anonymization protocols we adhered to are crucial for protecting client confidentiality. However, this necessary step might slightly diminish the specificity of the counseling dialogues, potentially impacting the LLMs' performance. Our evaluations suggest a performance reduction of approximately 6\% due to anonymization, as shown in Tab.~\ref{tab:role_ablation_anonymous}. Future research could explore advanced data protection techniques that preserve client privacy without significantly compromising model performance, such as federated learning.

\paragraph{Resource Constraints} Given the constraints of our budget and computational resources, we were limited to only evaluating 21 cutting-edge LLMs, as detailed in Tab.~\ref{tab:model_ablation}. While these evaluations provide valuable insights, further assessments of newer models are essential for practical applications. Besides, natively employing the largest model, Qwen1.5-110B-Chat, is computationally intensive, necessitating substantial resources, and we offer our fine-tuned model as a more efficient alternative with greater effectiveness.

\paragraph{Lack of Existing Benchmarks} As the pioneering study to utilize LLMs for predicting OCEAN traits from counseling dialogues, our experiments underscores the novelty and innovation of our framework. Despite our extensive efforts to validate the framework and explore its broader implications, the lack of pre-existing benchmarks or comparable studies necessitated the independent development of our experimental and evaluation methodologies. Creating standardized evaluation metrics and benchmarks would significantly enhance cross-study comparisons and drive further advancements.

\section*{Acknowledgement}
We express our gratitude to Yu Lu, Lili Pan, Lichao Zhang, and Qiyue Kang for their assistance with proofreading and valuable discussions. We also wish to acknowledge the support from the Google Developer Relations Team for providing GCP credits, and the Google Gemini API Team for granting access to the Gemini-ultra API. This research is funded by Westlake University Research Center for Industries of the Future (Grant No.WU2023C017) and Hangzhou Postdoctoral Research Fundings for Lizhi Ma (103110316582201) and Jingsong Ma (103110316582102, 103110316582101)

\newpage

\bibliography{custom}

\newpage

\appendix

\section{Appendices}

\subsection{Psychological Questionnaire}
\label{sec:appendix:questionnaire}

\subsubsection{BFI-2}

The items from original BFI-2 are as follows:

\noindent \fbox{%
    \begin{minipage}{\dimexpr\linewidth-2\fboxsep-2\fboxrule\relax}
        \small
        I am someone who ...
        \begin{enumerate}
            \setlength{\itemsep}{0pt}
            \setlength{\parsep}{0pt}
            \setlength{\parskip}{0pt}
            \item Is outgoing, sociable.
            \item Is compassionate, has a soft heart.
            \item Tends to be disorganized.
            \item Is relaxed, handles stress well.
            \item Has few artistic interests.
            \item Has an assertive personality.
            \item Is respectful, treats others with respect.
            \item Tends to be lazy.
            \item Stays optimistic after experiencing a setback.
            \item Is curious about many different things.
            \item Rarely feels excited or eager.
            \item Tends to find fault with others.
            \item Is dependable, steady.
            \item Is moody, has up and down mood swings.
            \item Is inventive, finds clever ways to do things.
            \item Tends to be quiet.
            \item Feels little sympathy for others.
            \item Is systematic, likes to keep things in order.
            \item Can be tense.
            \item Is fascinated by art, music, or literature.
            \item Is dominant, acts as a leader.
            \item Starts arguments with others.
            \item Has difficulty getting started on tasks.
            \item Feels secure, comfortable with self.
            \item Avoids intellectual, philosophical discussions.
            \item Is less active than other people.
            \item Has a forgiving nature.
            \item Can be somewhat careless.
            \item Is emotionally stable, not easily upset.
            \item Has little creativity.
            \item Is sometimes shy, introverted.
            \item Is helpful and unselfish with others.
            \item Keeps things neat and tidy.
            \item Worries a lot.
            \item Values art and beauty.
            \item Finds it hard to influence people.
            \item Is sometimes rude to others.
            \item Is efficient, gets things done.
            \item Often feels sad.
            \item Is complex, a deep thinker.
            \item Is full of energy.
            \item Is suspicious of others’ intentions.
            \item Is reliable, can always be counted on.
            \item Keeps their emotions under control.
            \item Has difficulty imagining things.
            \item Is talkative.
            \item Can be cold and uncaring.
            \item Leaves a mess, doesn’t clean up.
            \item Rarely feels anxious or afraid.
            \item Thinks poetry and plays are boring.
            \item Prefers to have others take charge.
            \item Is polite, courteous to others.
            \item Is persistent, works until the task is finished.
            \item Tends to feel depressed, blue.
            \item Has little interest in abstract ideas.
            \item Shows a lot of enthusiasm.
            \item Assumes the best about people.
            \item Sometimes behaves irresponsibly.
            \item Is temperamental, gets emotional easily.
            \item Is original, comes up with new ideas.
        \end{enumerate}
    \end{minipage}
}

\begin{table*}[t]
    \centering
    \resizebox{\textwidth}{!}{
        \begin{tabular}{lr|rrrrrr}
            \toprule
                                                           & Cronbach $\alpha$ & Extraversion & Agreeableness & Conscientiousness & Negative Emotionality & Open Mindedness & Kappa Avg. \\
            Model                                          &                   &              &               &                   &                       &                 &            \\
            \midrule
            gemini-1.0-pro-001~\cite{geminiteam2024gemini} & 0.839             & 0.526        & 0.479         & 0.512             & 0.546                 & 0.426           & 0.498      \\
            \midrule
            Qwen1.5-110B-Chat~\cite{bai2023qwen}           & 0.814             & 0.711        & 0.233         & 0.678             & 0.630                 & 0.572           & 0.565      \\
            Qwen-72B-Chat~\cite{bai2023qwen}               & 0.776             & 0.428        & 0.432         & 0.457             & 0.501                 & 0.305           & 0.425      \\
            Meta-Llama-3-70B-Instruct~\cite{llama3}        & 0.808             & 0.758        & 0.635         & 0.671             & 0.888                 & 0.668           & 0.724      \\
            Yi-34B-Chat~\cite{ai2024yi}                    & 0.792             & -0.004       & -0.002        & -0.005            & 0.078                 & -0.002          & 0.013      \\
            AquilaChat2-34B~\cite{Aquila2}                 & 0.499             & 0.125        & 0.083         & 0.079             & 0.069                 & 0.082           & 0.088      \\
            internlm2-chat-20b~\cite{cai2024internlm2}     & 0.693             & 0.374        & 0.210         & 0.297             & 0.133                 & 0.230           & 0.249      \\
            Baichuan2-13B-Chat~\cite{yang2023baichuan}     & 0.771             & 0.442        & 0.343         & 0.376             & 0.445                 & 0.378           & 0.397      \\
            chatglm3-6b-128k~\cite{zeng2023glmb}           & 0.807             & 0.293        & 0.296         & 0.301             & 0.255                 & 0.275           & 0.284      \\
            \midrule
            Llama-3-8b-BFI(Ours)                           & 0.708             & 0.435        & 0.405         & 0.317             & 0.499                 & 0.373           & 0.406      \\
            \bottomrule
        \end{tabular}
    }
    \caption{\small \textbf{Internal consistency and test-retest reliability of LLMs in OCEAN traits prediciton task.}}
    \label{tab:reliability_models}
\end{table*}

The BFI-2 consists of 60 items, with each set of 12 items representing one of the five traits: Extraversion, Agreeableness, Conscientiousness, Negative Emotionality, and Open Mindedness. Participants rate their agreement with each statement on a 5-point Likert scale: 1. Disagree Strongly, 2. Disagree a Little, 3. Neutral, 4. Agree a Little, 5. Agree Strongly. Trait scores are determined by summing the scores of the relevant items from BFI Scoring system~\cite{soto2017next}, with higher scores reflecting higher levels of the trait.

In our research, we utilized the Chinese adaptation of the Big Five Inventory-2 (BFI-2)~\cite{soto2017next} to evaluate OCEAN traits. Items were embedded into the prompt template described in Sec.~\ref{sec:method:prompt}, and the LLMs produced responses as answers to the questionnaire. We selected the BFI-2 due to its proven reliability and validity in assessing personality traits. Unlike the MBTI, which was utilized in some earlier studies, we elaborate on the differences and our rationale for this choice in the subsequent section.

\subsubsection{MBTI Questionnaire}

The Myers-Briggs Type Indicator (MBTI)~\cite{myers1962myers} is another widely used tool for personality assessment, based on Carl Jung's theory of psychological types. The MBTI categorizes individuals into one of 16 personality types based on four dichotomies: Extraversion (E) vs. Introversion (I), Sensing (S) vs. Intuition (N), Thinking (T) vs. Feeling (F), and Judging (J) vs. Perceiving (P). Each individual is assigned a four-letter type based on their preferences in each dichotomy.

Although MBTI is popular and widely used, the validity and reliability of MBTI have been questioned by the psychological community. There are three main criticisms of the MBTI compared to the BFI: (1) lack of scientific validity and reliability: the MBTI has been criticized for its lack of empirical support and scientific rigor~\cite{Diekmann2016FindingTR}. (2) binary nature and lack of nuance: the MBTI's type-based approach forces individuals into one of 16 types, which can oversimplify the complexity of human personality, while BFI measures personality across five dimensions, allowing for a more nuanced understanding~\cite{sava2011personality,Diekmann2016FindingTR}. (3) limited predictive power and practical application: the MBTI has been found to have limited predictive power regarding behavior and job performance, while the BFI has demonstrated better predictive validity in various contexts~\cite{furnham2015personality,Diekmann2016FindingTR,silpa2023robust}.

In conclusion, these factors limit the utility of the MBTI compared to the BFI, making the BFI a more robust and scientifically supported tool for personality assessment. With this consideration, we chose BFI in our study for better reliability and validity.

\subsection{Data Preprocessing Details}
\label{sec:appendix:data_preprocessing_details}

This section outlines the comprehensive data preprocessing steps undertaken to ready the counseling dialogues for training the LLMs. The preprocessing pipeline includes several crucial stages: 1. Data Collection, 2. Data Cleaning, 3. Anonymization, 4. Template Generation, and 5. Tokenization.

\paragraph{Data Collection:} Utilizing our counseling platform, we initiated our research through this medium. We gathered 853 counseling sessions from the platform, each consisting of a dialogue between a counselor and a client. These sessions were conducted in Chinese and spanned various subjects, such as mental health, relationships, and personal development. Participants were notified that their conversations would be used for research and gave their consent for their data to be included in this study.

\paragraph{Data Cleaning:} We conducted thorough data cleaning to eliminate any illegal characters and extraneous information from the counseling dialogues. This step was essential to maintain the quality and integrity of the data for OCEAN trait prediction.

\paragraph{Anonymization:} To safeguard the privacy and confidentiality of the participants, we anonymized 242 counseling dialogues by eliminating any personally identifiable information, including names, locations, and specific details that could disclose the participants' identities. This anonymization was crucial to guarantee the ethical utilization of the data in our research.

\paragraph{Template Creation:} We developed multiple prompt templates to simulate counseling conversations between a counselor and a client, as detailed in Sec.~\ref{sec:method:prompt} and Sec.~\ref{sec:appendix:prompt_lists}. These templates facilitated the generation of responses to the BFI-2 from the counseling dialogues, allowing the LLMs to infer the OCEAN traits.

\paragraph{Tokenization:} We tokenized the counseling dialogues following the corresponding tokenizer offered by the LLMs. The dialogue text was applied to chat template from the tokenizer, keep consistency with the instructional fine-tuning process.

\subsection{Prompts Used in Our Framework}
\label{sec:appendix:prompt_lists}

As discussed in Sec.~\ref{sec:method:prompt}, we introduce the prompt templates for the roles of ``counselor'' and ``observer'' utilized in our study to generate responses for the BFI-2.

\subsubsection{Counselor}

\noindent \fbox{%
    \begin{minipage}{\dimexpr\linewidth-2\fboxsep-2\fboxrule\relax}
        \small
        \textbf{System Prompt:} Act like a real counselor and do not mention anything with AI. You are a professional psychological counselor, and you are about to participate in a psycho-counseling. \\
        --- \\
        \textbf{User:} \{utterance 1 from client\} \\
        \textbf{LLM:} \{utterance 1 from counselor\} \\
        \textbf{User:} \{utterance 2 from client\} \\
        \textbf{LLM:} \{utterance 2 from counselor\} \\
        ... \\
        \textbf{User:} Before we end today's counseling session, please complete the following questionnaire based on the conversation and client's situation: \\
        --- \\
        \textbf{Question:} \{item from BFI\} \\
        \textbf{Options:} \\
        \textit{1. Disagree (strongly)} \\
        \textit{2. Disagree (a little)} \\
        \textit{3. Neutral (no opinion)} \\
        \textit{4. Agree (a little)} \\
        \textit{5. Agree (strongly)} \\
        --- \\
        Please tell me your choice and explain the reason:
    \end{minipage}%
}

\subsubsection{Observer}

\noindent \fbox{%
    \begin{minipage}{\dimexpr\linewidth-2\fboxsep-2\fboxrule\relax}
        \small
        \textbf{System Prompt:} You are an AI proficient in dialogue analysis and character profiling. Your task is to help the counselor analyze the utterance of the counseling dialogue. You need to answer a series of questions about the client's OCEAN traits based on the information in the chat records. \\
        --- \\
        Here come the dialogue: \\
        \textbf{User:} \{utterance 1 from client\} \\
        \textbf{Counselor:} \{utterance 1 from counselor\} \\
        \textbf{User:} \{utterance 2 from client\} \\
        \textbf{Counselor:} \{utterance 2 from counselor\} \\
        ... \\
        --- \\
        Based on the dialogue, please provide the most appropriate option for the following question: \\
        \textbf{Question:} \{item from BFI\} \\
        \textbf{Options:} \\
        \textit{1. Disagree (strongly)} \\
        \textit{2. Disagree (a little)} \\
        \textit{3. Neutral (no opinion)} \\
        \textit{4. Agree (a little)} \\
        \textit{5. Agree (strongly)} \\
        --- \\
        Please tell me your choice and explain the reason:
    \end{minipage}%
}

\input{tables/statistics_dialogues.tex}

\subsection{Reliability Evaluation}
\label{sec:appendix:reliability}

To ensure the robustness and applicability of our proposed method, we adopt a comprehensive suite of metrics aimed at evaluating both the validity and reliability of LLMs in predicting OCEAN traits. This section delineates the specific metrics employed in our study, underscoring their significance in psychological evaluation.

\subsubsection{Reliability Metrics}
Reliability, in the context of psychological assessments, denotes the consistency and stability of a test across multiple administrations. A reliable test consistently reflects the true psychological characteristic it aims to measure, rather than being influenced by random error or variability. This concept is paramount in our evaluation to ascertain that the LLMs are not merely "Stochastic Parrots" but are genuinely reflective of the OCEAN traits. We utilize two primary metrics to assess reliability.

\noindent 1.\textbf{Internal Consistency:} This metric evaluates the degree of correlation among individual test items, ensuring that they collectively measure the same construct. We employ Cronbach's Alpha ($\alpha$) as the statistical measure for internal consistency. A higher $\alpha$ value indicates a more reliable construct measurement, with values above 0.7 generally considered acceptable in psychological research.

\noindent 2.\textbf{Test-Retest Reliability:} To measure the stability of our method over time, we apply the Kappa statistic, which assesses the consistency of test results upon repeated administrations under similar conditions. A higher Kappa value suggests greater reliability, indicating that the LLMs' predictions of the OCEAN traits are stable over time.

\begin{table}[ht]
    \centering
    \resizebox{0.45\textwidth}{!}{
        \begin{tabular}{lrrrrrr}
            \toprule
                   & O     & C     & E     & A     & N     & Avg.  \\
            Try \# &       &       &       &       &       &       \\
            \midrule
            0      & 0.660 & 0.650 & 0.577 & 0.401 & 0.636 & 0.585 \\
            1      & 0.658 & 0.609 & 0.593 & 0.375 & 0.587 & 0.564 \\
            2      & 0.697 & 0.638 & 0.612 & 0.413 & 0.579 & 0.588 \\
            3      & 0.646 & 0.650 & 0.629 & 0.416 & 0.618 & 0.592 \\
            4      & 0.636 & 0.592 & 0.597 & 0.425 & 0.632 & 0.576 \\
            5      & 0.670 & 0.662 & 0.567 & 0.397 & 0.610 & 0.581 \\
            6      & 0.646 & 0.627 & 0.555 & 0.407 & 0.617 & 0.570 \\
            7      & 0.657 & 0.618 & 0.617 & 0.367 & 0.644 & 0.581 \\
            8      & 0.680 & 0.641 & 0.647 & 0.386 & 0.600 & 0.591 \\
            9      & 0.630 & 0.648 & 0.585 & 0.417 & 0.621 & 0.580 \\
            Avg.   & 0.658 & 0.633 & 0.598 & 0.400 & 0.614 & 0.581 \\
            Std.   & 0.019 & 0.021 & 0.027 & 0.018 & 0.020 & 0.008 \\
            \bottomrule
        \end{tabular}
    }
    \caption{\small \textbf{PCC of 10 tries for test-retest reliability of Llama3-8B model.}}
    \label{tab:reliability_test-retest}
\end{table}

Using these meticulously chosen metrics, our study aims to rigorously evaluate and validate the ability of LLMs to accurately predict OCEAN traits based on counseling dialogues. The subsequent sections will elaborate on our innovative approach to simulating counseling interactions and detail the methodology employed to ensure the accuracy and reliability of our predictions.

\subsection{Ablation Study}

\subsubsection{Performance Drop in Anonymization}

Privacy and confidentiality are paramount in counseling sessions, which requires anonymization of client data. However, this anonymization process can inadvertently affect the performance of LLMs in predicting OCEAN traits. To quantify this impact, we performed an ablation study to evaluate the performance drop due to anonymization. We compared the PCC of Qwen1.5-110B-Chat and our Llama-3-8b-BFI to predict OCEAN traits with and without anonymization, as shown in Tab.~\ref{tab:role_ablation_anonymous}.

\begin{table*}[t]
    \centering
    \resizebox{\textwidth}{!}{
        \begin{tabular}{llllllllr}
            \toprule
                                                  &                               &           & Open Mindedness & Conscientiousness & Extraversion & Agreeableness & Negative Emotionality & Avg.  \\
            Model                                 & Role                          & Anonymous &                 &                   &              &               &                       &       \\
            \midrule
            \multirow[t]{6}{*}{Qwen1.5-110B-Chat} & \multirow[t]{2}{*}{client}    & False     & 0.455***        & 0.463***          & 0.521***     & 0.334**       & 0.354**               & 0.425 \\
                                                  &                               & True      & 0.401***        & 0.482***          & 0.483***     & 0.256*        & 0.352**               & 0.395 \\
            \cline{2-9}
                                                  & \multirow[t]{2}{*}{counselor} & False     & 0.314**         & 0.354**           & 0.488***     & 0.050         & 0.422***              & 0.326 \\
                                                  &                               & True      & 0.328**         & 0.357**           & 0.455***     & 0.039         & 0.395***              & 0.315 \\
            \cline{2-9}
                                                  & \multirow[t]{2}{*}{observer}  & False     & 0.375**         & 0.341**           & 0.436***     & 0.378**       & 0.400***              & 0.386 \\
                                                  &                               & True      & 0.328**         & 0.306*            & 0.416***     & 0.381**       & 0.370**               & 0.360 \\
            \cline{1-9} \cline{2-9}
            \multirow[t]{6}{*}{Llama-3-8b-BFI}    & \multirow[t]{2}{*}{client}    & False     & 0.694***        & 0.653***          & 0.625***     & 0.524***      & 0.661***              & 0.631 \\
                                                  &                               & True      & 0.692***        & 0.554***          & 0.569***     & 0.448***      & 0.648***              & 0.582 \\
            \cline{2-9}
                                                  & \multirow[t]{2}{*}{counselor} & False     & 0.657***        & 0.621***          & 0.560***     & 0.361**       & 0.570***              & 0.554 \\
                                                  &                               & True      & 0.652***        & 0.586***          & 0.550***     & 0.412***      & 0.539***              & 0.548 \\
            \cline{2-9}
                                                  & \multirow[t]{2}{*}{observer}  & False     & 0.585***        & 0.518***          & 0.544***     & 0.484***      & 0.510***              & 0.528 \\
                                                  &                               & True      & 0.499***        & 0.560***          & 0.476***     & 0.357**       & 0.483***              & 0.475 \\
            \bottomrule
        \end{tabular}

    }
    \caption{\small \textbf{Ablation for performance drop when applying anonymization.}}
    \label{tab:role_ablation_anonymous}
\end{table*}

\begin{table*}
    \resizebox*{\textwidth}{!}{
        \begin{tabular}{lllllllr}
            \toprule
                                    &                   & Open Mindedness & Conscientiousness & Extraversion & Agreeableness & Negative Emotionality & Avg.  \\
            Granularity             & Model Name        &                 &                   &              &               &                       &       \\
            \midrule
            \multirow[t]{2}{*}{0.1} & Llama-3-8b-BFI    & 0.347**         & 0.269*            & 0.304*       & 0.341**       & 0.202                 & 0.293 \\
                                    & Qwen1.5-110B-Chat & 0.032           & 0.039             & 0.104        & 0.186         & 0.131                 & 0.098 \\
            \cline{1-8}
            \multirow[t]{2}{*}{0.2} & Llama-3-8b-BFI    & 0.558***        & 0.515***          & 0.366**      & 0.518***      & 0.409***              & 0.473 \\
                                    & Qwen1.5-110B-Chat & 0.184           & 0.372**           & 0.396***     & 0.365**       & 0.259*                & 0.315 \\
            \cline{1-8}
            \multirow[t]{2}{*}{0.3} & Llama-3-8b-BFI    & 0.664***        & 0.464***          & 0.506***     & 0.465***      & 0.452***              & 0.510 \\
                                    & Qwen1.5-110B-Chat & 0.337**         & 0.337**           & 0.378**      & 0.284*        & 0.317**               & 0.331 \\
            \cline{1-8}
            \multirow[t]{2}{*}{0.4} & Llama-3-8b-BFI    & 0.647***        & 0.546***          & 0.567***     & 0.455***      & 0.505***              & 0.544 \\
                                    & Qwen1.5-110B-Chat & 0.272*          & 0.456***          & 0.370**      & 0.320**       & 0.319**               & 0.347 \\
            \cline{1-8}
            \multirow[t]{2}{*}{0.5} & Llama-3-8b-BFI    & 0.723***        & 0.559***          & 0.536***     & 0.481***      & 0.520***              & 0.564 \\
                                    & Qwen1.5-110B-Chat & 0.401***        & 0.360**           & 0.350**      & 0.256*        & 0.310*                & 0.335 \\
            \cline{1-8}
            \multirow[t]{2}{*}{0.6} & Llama-3-8b-BFI    & 0.740***        & 0.628***          & 0.552***     & 0.470***      & 0.568***              & 0.592 \\
                                    & Qwen1.5-110B-Chat & 0.461***        & 0.410***          & 0.391***     & 0.372**       & 0.296*                & 0.386 \\
            \cline{1-8}
            \multirow[t]{2}{*}{0.7} & Llama-3-8b-BFI    & 0.715***        & 0.628***          & 0.614***     & 0.492***      & 0.598***              & 0.609 \\
                                    & Qwen1.5-110B-Chat & 0.370**         & 0.374**           & 0.381**      & 0.363**       & 0.303*                & 0.358 \\
            \cline{1-8}
            \multirow[t]{2}{*}{0.8} & Llama-3-8b-BFI    & 0.695***        & 0.650***          & 0.638***     & 0.505***      & 0.663***              & 0.630 \\
                                    & Qwen1.5-110B-Chat & 0.371**         & 0.509***          & 0.407***     & 0.351**       & 0.346**               & 0.397 \\
            \cline{1-8}
            \multirow[t]{2}{*}{0.9} & Llama-3-8b-BFI    & 0.709***        & 0.631***          & 0.648***     & 0.536***      & 0.632***              & 0.631 \\
                                    & Qwen1.5-110B-Chat & 0.371**         & 0.517***          & 0.438***     & 0.334**       & 0.296*                & 0.391 \\
            \cline{1-8}
            \multirow[t]{2}{*}{1.0} & Llama-3-8b-BFI    & 0.704***        & 0.609***          & 0.632***     & 0.443***      & 0.696***              & 0.617 \\
                                    & Qwen1.5-110B-Chat & 0.455***        & 0.463***          & 0.521***     & 0.334**       & 0.354**               & 0.425 \\
            \cline{1-8}
        \end{tabular}
    }
    \caption{\small \textbf{PCC of ablation for different granularity levels.}
        With the increase in granularity, the PCC values increase for both models, indicating that the granularity level significantly impacts the performance of LLMs in predicting OCEAN traits.
    }
    \label{tab:granularity}
\end{table*}

\begin{table*}[t]
    \centering
    \resizebox{0.9\linewidth}{!}{
        \begin{tabular}{llllllr}
            \toprule
                        & Open Mindedness & Conscientiousness & Extraversion & Agreeableness & Negative Emotionality & Avg.  \\
            Alignment   &                 &                   &              &               &                       &       \\
            \midrule
            DPO w/ SFT  & 0.692***        & 0.554***          & 0.569***     & 0.448***      & 0.648***              & 0.582 \\
            DPO w/o SFT & 0.655***        & 0.511***          & 0.592***     & 0.531***      & 0.527***              & 0.563 \\
            \bottomrule
        \end{tabular}
    }
    \caption{\small \textbf{PCC of w/ and w/o SFT in alignment.}
        The alignment process with SFT improves the performance of Llama3-8B model in predicting OCEAN traits.
    }
    \label{tab:alignment}
\end{table*}

\subsubsection{Ablation for Assigning Specific Roles in Role-Playing}

As mentioned in Sec.~\ref{sec:exp:rq1}, we explored the impact of various roles in the role-playing context. A pertinent question arises: ``Does the performance of LLMs change based on the specific roles assigned in the role-playing scenario?'' To investigate this, we performed an ablation study to assess how well LLMs predict OCEAN traits when particular roles are designated in the role-playing environment.

In a standard counseling scenario, the roles of ``Client'', ``Counselor'', and ``Observer'' are fundamental. We assigned ten renowned psychologists to the roles of ``Counselor'' or ``Observer'' to leverage their expertise for LLMs. For comparison purposes, we also included four common names and one name composed of random characters.

Unexpectedly, the findings in Tab.~\ref{tab:role_ablation_extend} indicate that assigning particular roles does not offer any extra advantage. When famous psychologists are assigned to LLM, the performance actually decreases compared to using common names and random characters. For the observer, the performance of famous psychologists is comparable to that of common names and random characters.

This contradicts our initial assumption, as our LLM does not gain from the conditioning of renowned psychologists, possibly due to the significant disparity between the actual counselor and the famous psychologists. This outcome implies that the optimal approach for our framework is to allocate the three inherent roles within the role-playing scenario.

\begin{table*}[t]
    \centering
    \resizebox{0.95\textwidth}{!}{
        \begin{tabular}{llllllr}
            \toprule
                                     & Open Mindedness & Conscientiousness & Extraversion & Agreeableness & Negative Emotionality & Avg.  \\
            Role                     &                 &                   &              &               &                       &       \\
            \midrule
            counselor                & 0.652***        & 0.586***          & 0.550***     & 0.412***      & 0.539***              & 0.548 \\
                                     & \cline{1-6}
            counselor-B.F. Skinner   & 0.570***        & 0.653***          & 0.596***     & 0.290*        & 0.560***              & 0.534 \\
            counselor-Ivan Pavlov    & 0.513***        & 0.568***          & 0.505***     & 0.304*        & 0.524***              & 0.483 \\
            counselor-Lev Vygotsky   & 0.560***        & 0.594***          & 0.594***     & 0.292*        & 0.561***              & 0.520 \\
            counselor-Carl Rogers    & 0.580***        & 0.560***          & 0.559***     & 0.178         & 0.536***              & 0.483 \\
            counselor-Harry Harlow   & 0.564***        & 0.580***          & 0.519***     & 0.283*        & 0.518***              & 0.493 \\
            counselor-William James  & 0.522***        & 0.509***          & 0.528***     & 0.418***      & 0.514***              & 0.498 \\
            counselor-Anna Freud     & 0.583***        & 0.452***          & 0.629***     & 0.352**       & 0.476***              & 0.498 \\
            counselor-Sigmund Freud  & 0.461***        & 0.541***          & 0.576***     & 0.291*        & 0.628***              & 0.499 \\
            counselor-Jean Piaget    & 0.522***        & 0.563***          & 0.593***     & 0.186         & 0.511***              & 0.475 \\
            counselor-Albert Bandura & 0.558***        & 0.615***          & 0.506***     & 0.291*        & 0.512***              & 0.496 \\
            Avg.                     &                 &                   &              &               &                       & 0.497 \\
                                     & \cline{1-6}
            counselor-Zhang3         & 0.627***        & 0.645***          & 0.498***     & 0.397***      & 0.495***              & 0.532 \\
            counselor-Li4            & 0.642***        & 0.548***          & 0.526***     & 0.457***      & 0.568***              & 0.548 \\
            counselor-Wang5          & 0.620***        & 0.599***          & 0.548***     & 0.286*        & 0.529***              & 0.516 \\
            counselor-Zhao6          & 0.664***        & 0.571***          & 0.587***     & 0.456***      & 0.522***              & 0.560 \\
            Avg.                     &                 &                   &              &               &                       & 0.539 \\
                                     & \cline{1-6}
            counselor-XXXX           & 0.657***        & 0.566***          & 0.654***     & 0.461***      & 0.554***              & 0.578 \\
            \midrule
            observer                 & 0.499***        & 0.560***          & 0.476***     & 0.357**       & 0.483***              & 0.475 \\
                                     & \cline{1-6}
            observer-B.F. Skinner    & 0.552***        & 0.532***          & 0.444***     & 0.216         & 0.526***              & 0.454 \\
            observer-Ivan Pavlov     & 0.484***        & 0.572***          & 0.512***     & 0.389**       & 0.472***              & 0.486 \\
            observer-Lev Vygotsky    & 0.640***        & 0.578***          & 0.502***     & 0.376**       & 0.511***              & 0.521 \\
            observer-Carl Rogers     & 0.531***        & 0.591***          & 0.415***     & 0.289*        & 0.545***              & 0.474 \\
            observer-Harry Harlow    & 0.506***        & 0.647***          & 0.456***     & 0.316**       & 0.490***              & 0.483 \\
            observer-William James   & 0.506***        & 0.534***          & 0.571***     & 0.314**       & 0.471***              & 0.479 \\
            observer-Anna Freud      & 0.616***        & 0.470***          & 0.489***     & 0.313**       & 0.531***              & 0.484 \\
            observer-Sigmund Freud   & 0.555***        & 0.523***          & 0.403***     & 0.322**       & 0.487***              & 0.458 \\
            observer-Jean Piaget     & 0.497***        & 0.577***          & 0.426***     & 0.287*        & 0.463***              & 0.450 \\
            observer-Albert Bandura  & 0.539***        & 0.613***          & 0.388**      & 0.319**       & 0.574***              & 0.487 \\
            Avg.                     &                 &                   &              &               &                       & 0.477 \\
                                     & \cline{1-6}
            observer-Zhang3          & 0.603***        & 0.690***          & 0.465***     & 0.325**       & 0.490***              & 0.515 \\
            observer-Li4             & 0.445***        & 0.486***          & 0.471***     & 0.349**       & 0.524***              & 0.455 \\
            observer-Wang5           & 0.443***        & 0.625***          & 0.489***     & 0.354**       & 0.444***              & 0.471 \\
            observer-Zhao6           & 0.445***        & 0.512***          & 0.499***     & 0.285*        & 0.608***              & 0.470 \\
            Avg.                     &                 &                   &              &               &                       & 0.477 \\
                                     & \cline{1-6}
            observer-XXXX            & 0.518***        & 0.511***          & 0.585***     & 0.308*        & 0.446***              & 0.474 \\
            \bottomrule
        \end{tabular}
    }
    \caption{\small \textbf{Effect of different roles on the performance of predicting OCEAN traits.}}
    \label{tab:role_ablation_extend}
\end{table*}

\subsubsection{Ablation for Different Models in Alignment}

We conducted an ablation study to evaluate the impact of different models in the alignment process. We employed the Qwen1.5-7B-Chat and Qwen2-7B-Instruct models to against the Meta-Llama-3-8B-Instruct model. Due to resource constraints, we only fine-tuned these models with 242 counseling dialogues and evaluated them on 611 dialogues. The results in Tab.~\ref{tab:fine_tuning_model_ablation} demonstrate that the fine-tuned models significantly outperform the original models across all OCEAN traits, indicating the effectiveness of the alignment process.

\begin{table*}[t]
    \centering
    \resizebox{\textwidth}{!}{
        \begin{tabular}{llllllllr}
            \toprule
                                                   & Train \# & Valid \# & Open Mindedness & Conscientiousness & Extraversion & Agreeableness & Negative Emotionality & Avg.  \\
            Model                                  &          &          &                 &                   &              &               &                       &       \\
            \midrule
            Meta-Llama-3-8B-Instruct~\cite{llama3} & -        & 242      & 0.177           & 0.434***          & 0.233        & 0.111         & 0.303*                & 0.252 \\
            Llama-3-8b-BFI (Ours)                  & 611      & 242      & 0.692***        & 0.554***          & 0.569***     & 0.448***      & 0.648***              & 0.582 \\
            \midrule
            Meta-Llama-3-8B-Instruct~\cite{llama3} & -        & 611      & 0.299**         & 0.255*            & 0.383***     & 0.080         & 0.337**               & 0.271 \\
            Llama-3-8b-BFI-242 (Ours)              & 242      & 611      & 0.566***        & 0.495***          & 0.538***     & 0.467***      & 0.512***              & 0.516 \\
            Qwen1.5-7B-Chat~\cite{bai2023qwen}     & -        & 611      & 0.266*          & 0.311**           & 0.274*       & 0.178         & 0.333**               & 0.272 \\
            Qwen1.5-7B-Chat-BFI-242 (Ours)         & 242      & 611      & 0.562***        & 0.470***          & 0.537***     & 0.378***      & 0.558***              & 0.501 \\
            Qwen2-7B-Instruct~\cite{bai2023qwen}   & -        & 611      & 0.280*          & 0.313**           & 0.305**      & 0.054         & 0.182                 & 0.227 \\
            Qwen2-7B-Instruct-BFI-242 (Ours)       & 242      & 611      & 0.502***        & 0.389***          & 0.502***     & 0.460***      & 0.557***              & 0.482 \\
            \bottomrule
        \end{tabular}
    }
    \caption{\small
        \textbf{PCC of ablation for different models in alignment.}
        ``Llama-3-8b-BFI-242'', ``Qwen1.5-7B-Chat-BFI-242'', and ``Qwen2-7B-Instruct-BFI-242'' denote the models fine-tuned with 242 counseling dialogues and evaluated on 611 dialogues. Compared to the original models, all fine-tuned models benefit from the alignment process, achieving higher and significant PCC values across all OCEAN traits.
    }
    \label{tab:fine_tuning_model_ablation}
\end{table*}

\subsection{Full OCEAN traits Prediction Correlation Results}
\label{sec:appendix:full_correlation_results}

In this section, we provide a comprehensive overview of the correlation outcomes for the OCEAN traits prediction. The results are categorized based on the primary LLMs employed in the experiments. The correlation outcomes are expressed as PCC between the predicted and actual OCEAN traits. PCC values span from -1 to 1, where 1 denotes a perfect positive linear relationship, -1 signifies a perfect negative linear relationship, and 0 represents the absence of a linear relationship between the predicted and actual OCEAN traits.

\subsubsection{Meta-Llama-3-8B-Instruct}

\begin{figure}[t]
    \centering
    \includegraphics[width=0.99\linewidth]{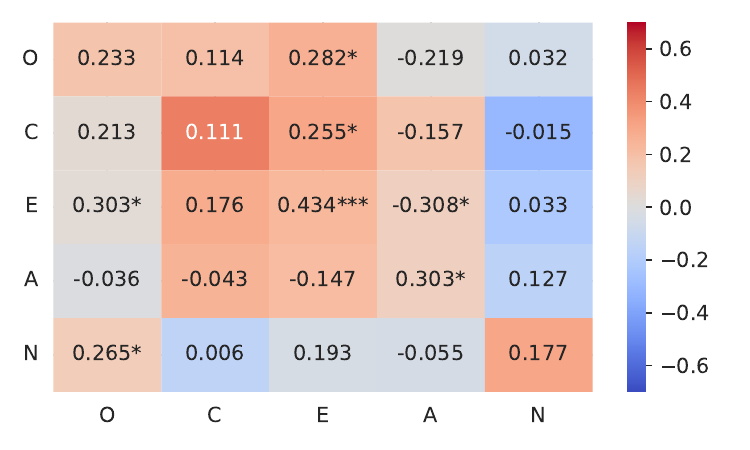}
    \caption{\small \textbf{PCC between predicted and actual OCEAN traits using Meta-Llama-3-8B-Instruct~\cite{llama3}.}}
    \label{fig:corr:meta-llama-3-8b-instruct}
\end{figure}

"Meta-Llama-3-8B-Instruct"~\cite{llama3} is a LLM developed and refined by Meta, demonstrating robust performance across various NLP tasks. This model served as the foundational model for aligning our LLM to the OCEAN traits prediction task. The correlation outcomes are illustrated in Fig.~\ref{fig:corr:meta-llama-3-8b-instruct}.

\subsubsection{Llama-3-8b-BFI}

\begin{figure}[t]
    \centering
    \includegraphics[width=0.99\linewidth]{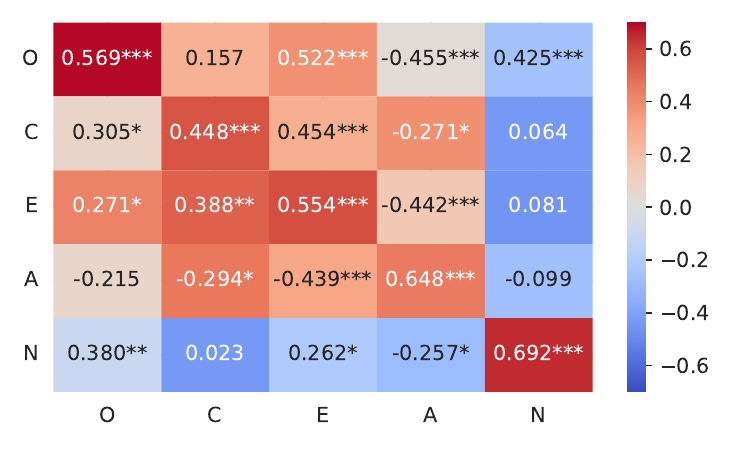}
    \caption{\small \textbf{PCC between predicted and actual OCEAN traits using Llama-3-8b-BFI (Ours).}}
    \label{fig:corr:Llama-3-8b-BFI}
\end{figure}

We adapted the Llama-3-8B model for the OCEAN traits prediction task and designated it as ``Llama-3-8b-BFI''. The correlation outcomes are illustrated in Fig.~\ref{fig:corr:Llama-3-8b-BFI}. This model attained the highest correlation as indicated in Tab.~\ref{tab:model_ablation}, providing a robust benchmark for the OCEAN traits prediction task.

\subsubsection{Qwen1.5-110B-Chat}

\begin{figure}[t]
    \centering
    \includegraphics[width=0.99\linewidth]{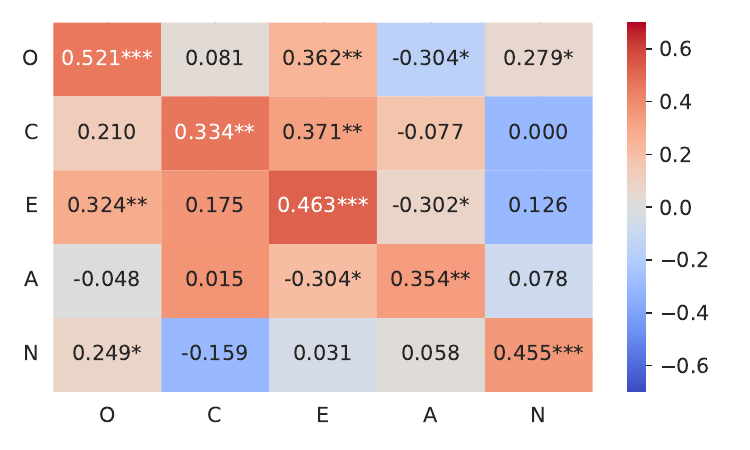}
    \caption{\small \textbf{PCC between predicted and actual OCEAN traits using qwen1.5-110b-chat~\cite{bai2023qwen}.}}
    \label{fig:corr:qwen1.5-110b-chat}
\end{figure}

``Qwen1.5-110B-Chat''~\cite{bai2023qwen} stands out as one of the most advanced and extensive LLMs available in the open-source domain. Its robust performance and inherent support for Chinese make it highly suitable for predicting OCEAN traits in Chinese counseling contexts. Achieving the highest correlation among open-source models, the correlation results are depicted in Fig.~\ref{fig:corr:qwen1.5-110b-chat}.

\subsubsection{DeepSeek-Chat}

\begin{figure}[t]
    \centering
    \includegraphics[width=0.99\linewidth]{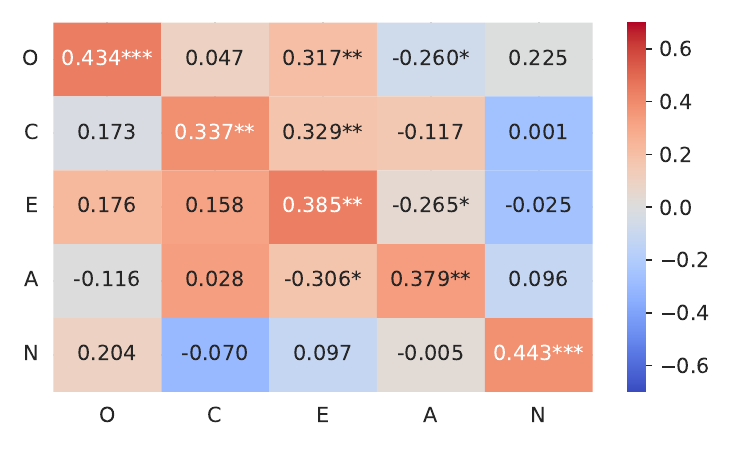}
    \caption{\small \textbf{PCC between predicted and actual OCEAN traits using deepseek-chat~\cite{deepseekai2024deepseekv2}.}}
    \label{fig:corr:deepseek-chat}
\end{figure}

"DeepSeek-Chat"~\cite{deepseekai2024deepseekv2} is an advanced LLM created by DeepSeek AI, and it is claimed to rival GPT4. We selected "DeepSeek-Chat" for multiple ablation studies in \ref{sec:exp:rq1} due to its excellent performance and affordable cost. The related correlation results are presented in Fig.~\ref{fig:corr:deepseek-chat}.

\subsubsection{Gemini-1.5-Pro}

\begin{figure}[t]
    \centering
    \includegraphics[width=0.99\linewidth]{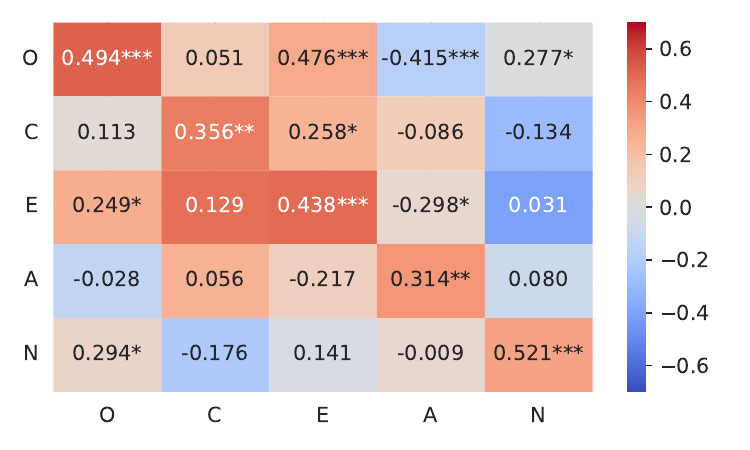}
    \caption{\small \textbf{PCC between predicted and actual OCEAN traits using Gemini-1.5-Pro~\cite{geminiteam2024gemini}.}}
    \label{fig:corr:gemini-1.5-pro}
\end{figure}

"Gemini-1.5-Pro"~\cite{geminiteam2024gemini} is a LLM developed by Google, featuring enhanced performance and abilities compared to its predecessor, Gemini-1.0 Pro, which utilizes a Mixture of Experts (MoE) architecture. The complete correlation results for its top performance among proprietary language models are presented in Fig.~\ref{fig:corr:gemini-1.5-pro}.

\subsubsection{GPT-4-Turbo}

\begin{figure}[t]
    \centering
    \includegraphics[width=0.99\linewidth]{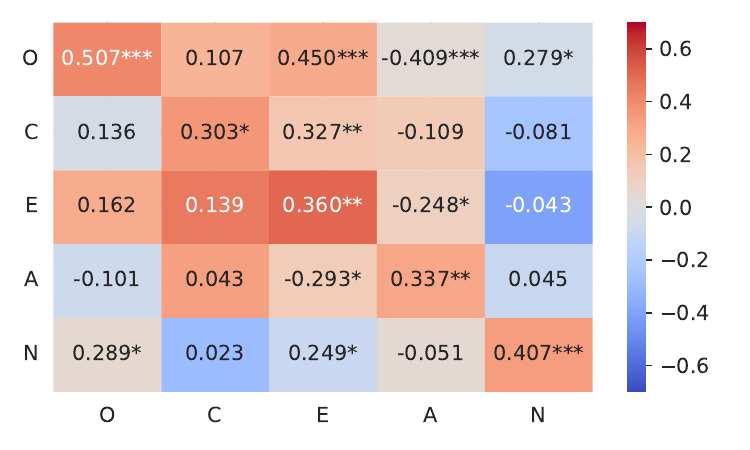}
    \caption{\small \textbf{PCC between predicted and actual OCEAN traits using GPT-4-Turbo~\cite{openai2023gpt4}.}}
    \label{fig:corr:gpt-4-turbo}
\end{figure}

Recognized as one of the most potent and widely utilized LLMs, ``GPT-4-Turbo''~\cite{openai2023gpt4} serves as a robust benchmark for predicting OCEAN traits. The correlation outcomes are illustrated in Fig.~\ref{fig:corr:gpt-4-turbo}.

\subsection{Overview of Hyper-Parameters}
\label{sec:appendix:hyperparameters}

The hyperparameters employed in our experiments are essential for ensuring the reproducibility and optimization of the Llama3-8B model in predicting Big Five Inventory traits. Below, we provide a comprehensive overview of the key hyperparameters, along with their descriptions and values, to offer a thorough understanding of the experimental configuration.

\begin{table*}
    \centering
    \resizebox{0.9\linewidth}{!}{
        \begin{tabular}{l|l|p{8cm}}
            \hline
            \textbf{Hyperparameter}              & \textbf{Value}            & \textbf{Description}                              \\
            \hline
            \texttt{Seed}                        & 42                        & Random seed for reproducibility                   \\
            \hline
            \texttt{Optimizer}                   & \texttt{AdamW}            & Optimizer used for training                       \\
            \hline
            \texttt{Learning Rate}               & 1e-6                      & Learning rate for optimizer                       \\
            \hline
            \texttt{Train Epochs \#}             & 3                         & Number of training epochs                         \\
            \hline
            \texttt{GPU \#}                      & 4 * Nvidia A100-SXM4-80GB & Number of GPUs                                    \\
            \hline
            \texttt{Per-device Train Batchsize}  & 1                         & Batch size per device during training             \\
            \hline
            \texttt{Gradient Accumulation Steps} & 2                         & Number of gradient accumulation steps             \\
            \hline
            \texttt{Warmup Ratio}                & 0.1                       & Ratio of warmup steps for learning rate scheduler \\
            \hline
            \texttt{LR Scheduler Type}           & \texttt{cosine}           & Learning rate scheduler type                      \\
            \hline
            \texttt{Data Type}                   & \texttt{bfloat16}         & Use bfloat16 precision during training            \\
            \hline
        \end{tabular}
    }
    \caption{\small \textbf{Key Hyperparameters for Fine-tuning LLM}}
    \label{tab:hyperparameters}
\end{table*}

Tab.~\ref{tab:hyperparameters} presents a summary of the key hyperparameters employed in our fine-tuning experiments. Each parameter is detailed to guarantee the clarity and reproducibility of our approach. This setup underscores our dedication to thorough and transparent research practices.

\end{document}

%% file: tables/statistics_dialogues.tex
\begin{table}[t]
    \centering
    \resizebox{\linewidth}{!}{
        \begin{tabular}{lrrr}
            \toprule
                                         & Total  & Counselor & Client \\
            \midrule
            \# Avg. sessions per speaker & -      & 95.44     & 10.48  \\
            \# Utterances                & 65,347 & 32,860    & 32,487 \\
            Avg. utterances per dialogue & 76.07  & 38.25     & 37.82  \\
            Avg. length per utterance    & 26.84  & 24.01     & 29.7   \\
            \bottomrule
        \end{tabular}
    }
    \caption{\small \textbf{Statistics of counseling dialogues from our platform.}}
    \label{tab:statistics_dialogues}
\end{table} 